\documentclass[10pt,twocolumn,letterpaper]{article}

\PassOptionsToPackage{table}{xcolor}

\usepackage[accept]{iccv}      % To produce the REVIEW version
\usepackage{microtype}
\usepackage{graphicx}
\usepackage{booktabs} % for professional tables
\usepackage{pifont}
\usepackage{multirow} 
\usepackage[most]{tcolorbox}

\tcbset{
  colframe=teal,        
  colback=green!20,     
  boxrule=0.3mm,       
  arc=3mm,             
  left=3mm,            
  right=3mm,           
  top=2mm,            
  bottom=2mm,         
}
\usepackage{colortbl}
\usepackage{xcolor}
\usepackage{wrapfig}
\usepackage{hyperref}

\usepackage[normalem]{ulem}
\usepackage{amsmath}
\usepackage{amssymb}
\usepackage{mathtools}
\usepackage{amsthm}

\usepackage[capitalize,noabbrev]{cleveref}

\theoremstyle{plain}
\newtheorem{theorem}{Theorem}[section]

\theoremstyle{definition}
\newtheorem{definition}[theorem]{Definition}
\newtheorem{assumption}[theorem]{Assumption}
\theoremstyle{remark}

\usepackage{microtype}
\definecolor{iccvblue}{rgb}{0.21,0.49,0.74}
\definecolor{lightpurple}{RGB}{232, 229, 250}
\def\paperID{8927} % *** Enter the Paper ID here
\def\confName{ICCV}
\def\confYear{2025}

\title{Incomplete In-context Learning}

\author{Wenqiang Wang\\
Sun Yat-sen University\\
{\tt\small wangwq69@mail2.sysu.edu.cn}
\and
Yangshijie Zhang\\
Lanzhou University\\
{\tt\small zhangyshj2023@lzu.edu.cn}
}

\begin{document}
\maketitle
\begin{abstract}
Large vision language models (LVLMs) achieve remarkable performance through Vision In-context Learning (VICL), a process that depends significantly on demonstrations retrieved from an extensive collection of annotated examples (retrieval database). Existing studies often assume that the retrieval database contains annotated examples for all labels. However, in real-world scenarios, delays in database updates or incomplete data annotation may result in the retrieval database containing labeled samples for only a subset of classes. We refer to this phenomenon as an \textbf{incomplete retrieval database} and define the in-context learning under this condition as \textbf{Incomplete In-context Learning (IICL)}. To address this challenge, we propose \textbf{Iterative Judgments and Integrated Prediction (IJIP)}, a two-stage framework designed to mitigate the limitations of IICL. The Iterative Judgments Stage reformulates an \(\boldsymbol{m}\)-class classification problem into a series of \(\boldsymbol{m}\) binary classification tasks, effectively converting the IICL setting into a standard VICL scenario. The Integrated Prediction Stage further refines the classification process by leveraging both the input image and the predictions from the Iterative Judgments Stage to enhance overall classification accuracy. IJIP demonstrates considerable performance across two LVLMs and two datasets under three distinct conditions of label incompleteness, achieving the highest accuracy of 93.9\%. Notably, even in scenarios where labels are fully available, IJIP still achieves the best performance of all six baselines. Furthermore, IJIP can be directly applied to \textbf{Prompt Learning} and is adaptable to the \textbf{text domain}.

\end{abstract}    
\section{Introduction}
\label{Introduction}
Large vision language models (LVLMs) demonstrate consistently high performance in downstream tasks through \emph{vision in-context learning} (VICL), which leverages demonstrations ~\citep{BrownICL,coda2023meta} retrieved from a retrieval database. Without explicitly modifying the LVLMs parameters, VICL reliably exceeds zero-shot prompt learning on diverse tasks~\citep{gonen-etal-2023-demystifying,gupta2023coverage}, such as image classification.
The success of VICL is largely based on the demonstrations retrieved from the retrieval database~\citep{kossen2024incontext,long2024decomposing,wang2023Latent,wu-etal-2023-self}.

\begin{figure}[t]
    \centering
    \begin{subfigure}[b]{\linewidth}
        \centering
        \includegraphics[width=0.85\textwidth]{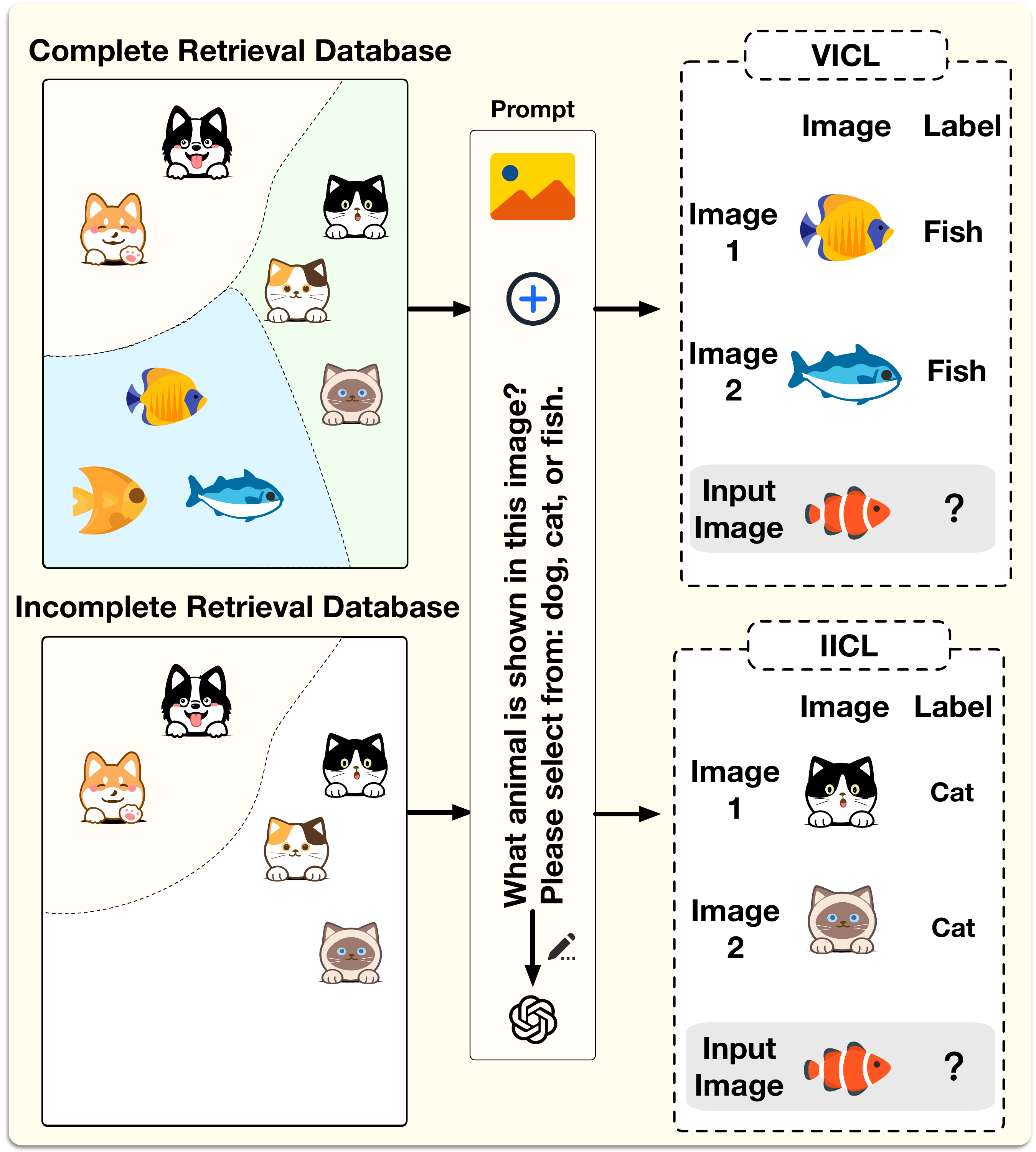}
        \caption{Comparison of complete vs incomplete retrieval databases and VICL vs IICL scenarios. In the \textit{incomplete retrieval database}, ``fish'' images are absent; thus, IICL fails to retrieve suitable demonstrations but retrieves ``cat'' images when the input is ``fish'', limiting IICL's performance. }       
        % no images labeled ``fish'' can be retrieved during the demonstration retrieval process. Consequently, when the input image depicts a ``fish'', \textit{IICL} can only retrieve images labeled ``cat'' as demonstrations

        \label{subfig_a1}  
    \end{subfigure}
    
    \vskip\baselineskip  
    
    \begin{subfigure}[b]{\linewidth}
        \centering
        \includegraphics[width=0.9\textwidth]{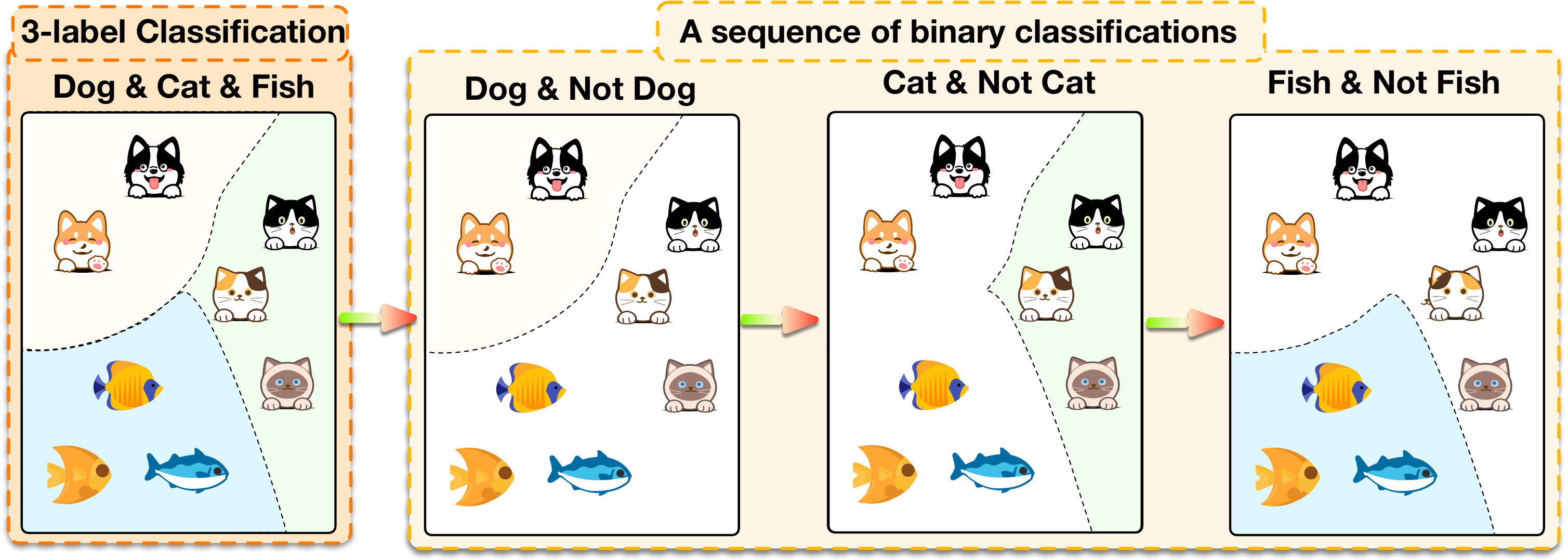}
        \caption{Comparison of $m$-label classification and a sequence of binary classifications. We transform a three-class classification task, ``What's this? Choose from dog, cat, and fish.'' into a sequence of binary classifications consisting of the following sub-questions based on the input image: (1) ``Is this a dog?'' (2) ``Is this a cat?'' and (3) ``Is this a fish?''.
 }  
        \label{subfig_b1}  
    \end{subfigure}

    \caption{The overview of incomplete retrieval database, IICL, and sequence of binary classifications.}
    \label{fig_comparison}
\end{figure}

\begin{figure*}[ht]
\centering
\includegraphics[width=0.9\textwidth]{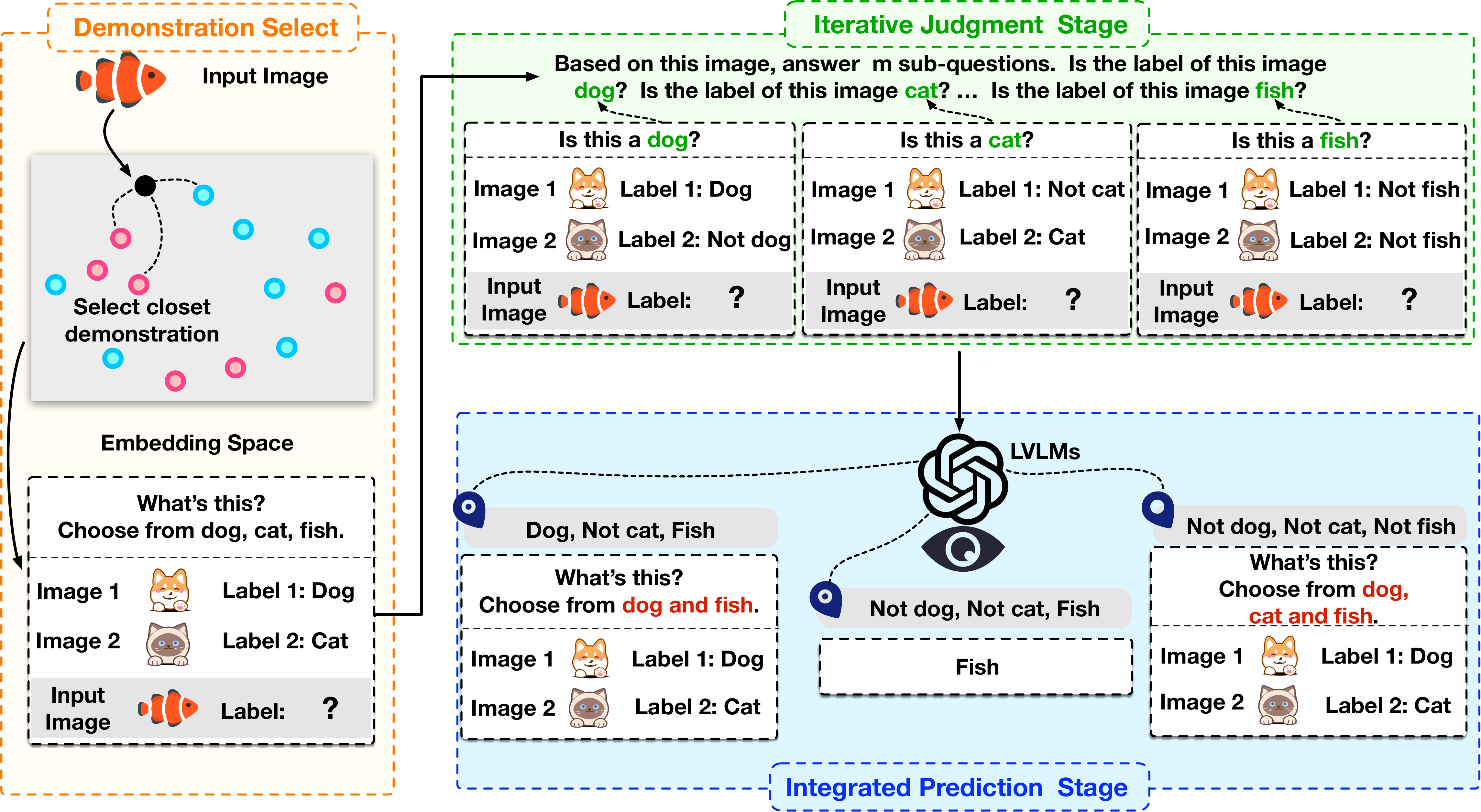}
\caption{
IJIP utilizes the CLIP model to vectorize images and retrieves the top 
$k$ most similar images from the incomplete retrieval database as demonstrations based on their similarity to the input image. In the Iterative Judgment Stage, IJIP first queries the LVLMs to classify the input image using the retrieved demonstrations. Specifically, IJIP identifies labels receiving positive responses as candidate classes. If no labels receive positive judgments, IJIP requests the LVLMs to perform a full-label classification. Conversely, if multiple labels receive positive judgments, IJIP initiates a second inquiry limited to these positively judged labels to refine the decision. Therefore, \textbf{IJIP queries LVLMs at most twice and at least once}, depending on the initial judgments.}
\label{overview}
\end{figure*}

Previous studies often assume that the retrieval database contains annotated examples for all labels. However, this assumption does not always hold in real-world scenarios. For example, new labels and their corresponding examples may emerge while the retrieval database is not updated promptly; in cases of extreme class imbalance, certain labels may have no available examples; or in Positive-Unlabeled (PU) learning, some labeled examples may remain unannotated. These limitations result in a retrieval database that contains only a subset of the annotated data corresponding to the full set of labels, as illustrated in Figure \ref{subfig_a1}. We define this phenomenon as the \textbf{incomplete retrieval database} and refer to the process of retrieving data from such a database for in-context learning as \textbf{Incomplete In-context Learning (IICL)}.

We perform an empirical study on IICL within the incomplete retrieval database scenario. Our results indicate that as the proportion of missing labels in the retrieval database increases, IICL performance progressively declines. Nevertheless, it consistently outperforms zero-shot prompting. Notably, even the absence of a single label leads to a substantial drop in performance. These findings underscore the importance of investigating IICL under incomplete retrieval database conditions.

To address the challenges posed by an incomplete retrieval database, we propose the \textit{Iterative Judgments and Integrated Prediction (IJIP)} method. As shown in Figure \ref{overview}, IJIP consists of two stages: the \textit{Iterative Judgments Stage} and the \textit{Integrated Prediction Stage}.
 In the \textit{Iterative Judgments Stage}, we assign additional labels to the images in the incomplete retrieval database. Specifically, if an image is labeled as \( C_j \), it is also implicitly not \( C_1, C_2, \dots, C_{j-1}, C_{j+1}, \dots, C_m \). This can be formally expressed as:
$\overline{C_1}, \overline{C_2}, \dots, \overline{C_{j-1}}, \overline{C_{j+1}}, \dots, \overline{C_m}.$
Therefore, we reformulate an \( m \)-class classification problem as a sequence of binary classifications, as illustrated in Figure \ref{subfig_b1},  where the sequence consists of \( m \) subproblems. In the $j$-th subproblem, the LVLMs determine whether the given image belongs to \( C_j \).  
To implement this, we modify the prompt in the LVLMs as follows:
\begin{quote}
\emph{``Based on this image, answer  \( m \) sub-questions.  Is the label of this image \( C_1 \)?  Is the label of this image \( C_2 \)? $\cdots$  Is the label of this image \( C_m \)?''}
\end{quote}

For the $j$-th subproblem, particularly the question \emph{``Is the label of this image \( C_j \)?''}, we assume that in the incomplete retrieval database, there exist \( w \) labels with available annotated data, where \( w < m \). In the original \( m \)-label classification setting, data for \( m-w \) labels is missing. However, in the transformed binary classification setting, no class is entirely absent. Specifically:
 If \( j \leq w \), both \( C_j \) and \( \overline{C_j} \) (non-\( C_j \)) examples exist in the database.
 If \( j > w \), at least \( \overline{C_j} \) examples are available.
By converting the IICL problem into a complete VICL problem through binary classification, the \textbf{Iterative Judgments Stage} effectively mitigates the challenge posed by missing labels in the retrieval database.

In the \textbf{Integrated Prediction Stage}, we refine the classification of the input image by leveraging both the image itself and the predictions obtained from the \textit{Iterative Judgments Stage}. The final decision-making process falls into one of three distinct scenarios.
\ding{182} All predictions are negative ($\overline{C_1}, \overline{C_2}, \dots, \overline{C_m}$). In this case, no class has been positively identified. We directly perform \textbf{\( m \)}-label VICL for classification.
\ding{183} Exactly one positive prediction is made (e.g., 
$\overline{C_1}, \overline{C_2}, \dots, C_j, \dots, \overline{C_m}$). The image is classified as \( C_j \), and we assign \( C_j \) as the final predicted label.
\ding{184} Multiple positive predictions are made (e.g., $\overline{C_1}, \overline{C_2}, \cdots, C_{j-1}, C_j, C_{j+1}, \cdots, \overline{C_m}$). Since multiple candidate labels exist, we perform an additional in-context learning classification among the positively predicted labels. For instance, if the predictions suggest the image could belong to $C_{j-1}, C_j$, or $C_{j+1}$, we refine the classification by treating it as a \textbf{three-class classification problem}.

This stage enhances classification accuracy by leveraging the predictions from the \textit{Iterative Judgments Stage}, thereby improving the final decision-making process.

We evaluate IJIP on two models and two datasets, achieving considerable performance with a peak accuracy of 93.9\%, exceeding the second-best method by 3.1\%. Even when the incomplete retrieval database contains data for only a single label, IJIP maintains a peak accuracy of 89.2\%.

 Notably, IJIP is adaptable to LVLMs of different sizes, achieving a maximum accuracy of 98.3\% across various LVLMs scales.
As a plug-and-play framework, IJIP is also applied to zero-shot prompt learning, where it further improves performance, yielding an average accuracy gain of 4.8\%. Additionally, even when the retrieval database contains data for all labels, IJIP achieves the SOTA  accuracy by 94.6\%.
Finally, IJIP is also effective in the text modality, outperforming the second-best method by 2.9\% in text classification tasks.
Our contributions are summarized as follows:
\begin{itemize}
    \item We introduce the concepts of the \textit{incomplete retrieval database} and \textit{Incomplete In-context Learning (IICL)} and conduct an empirical study to assess their impact. Our findings reveal that while IICL performance deteriorates as the number of missing labels increases, it consistently surpasses zero-shot prompting.
    
    \item We propose the Iterative Judgments and Integrated Prediction (IJIP) method, a \textit{framework} compatible with any Vision in-context learning (VICL) methods. 
   By reformulating the IICL problem into a standard VICL scenario, IJIP  effectively address the challenges posed by incomplete retrieval databases.

    \item IJIP achieves SOTA performance across multiple datasets and LVLMs. Furthermore, we successfully extend IICL to the \textit{text modality} and generalize IJIP to zero-shot prompt learning, demonstrating its versatility across different learning paradigms.
\end{itemize}
\label{Contribution}

\section{Incomplete In-context Learning}

In real-world scenarios, data is often dynamic, with new labels emerging gradually over time. 

 The retrieval database may not always be updated promptly to include data corresponding to newly introduced labels. For example, in \textit{class-incremental learning}, newly added labels may lack corresponding data in the model’s training dataset due to delayed updates.
Additionally, in extreme class-imbalance settings, certain classes may be underrepresented or entirely missing from the retrieval database. For instance, in \textit{extreme class-imbalance learning}, some classes may have no available data.
Moreover, in some cases, specific class-related data may remain unlabeled. For example, in \textit{positive and unlabeled learning}, only data associated with the positive label is annotated, while data corresponding to other labels remains unannotated.
We define the scenario in which the retrieval database contains only the data from the partial labels as \textit{incomplete retrieval database}. Let the set of labels be denoted as \( \{C_1, C_2, \cdots, C_m\} \). The incomplete retrieval database is formulated as follows:
\begin{equation}
\begin{array}{c}
D_{\text{in}} = \{ \mathbf{x}_i, y_i \}_{i=1}^n, \\
\textit{s.t. } y_i \in \{C_1, C_2, \cdots, C_w\}, \quad \text{where } w < m,
\end{array}
\end{equation}
where \( D_{\text{in}} \) denotes the incomplete retrieval database, which contains data only for labels \( \{C_1, C_2, \cdots, C_w\} \) and excludes data for labels \( \{C_{w+1}, C_{w+2}, \cdots, C_m\} \). 
We define VICL conducted under the incomplete retrieval database scenario as Incomplete In-context Learning (IICL).
The formal definition of IICL is provided in Section~\ref{Problem_Formulation}, and the impact of the incomplete retrieval database is examined in Section~\ref{Empirical_Study}.

\subsection{Problem Formulation}\label{Problem_Formulation}

\begin{definition}[\textbf{Incomplete In-context Learning (IICL)}]

VICL selects demonstrations from an incomplete retrieval database $D_{\text{in}}$, where the database contains data corresponding only to a subset of the labels, and certain labels lack associated data. It constructs the in-context with the
demonstrations retrieved from the incomplete retrieval database $D_{\text{in}}$ and predict the label \( \hat{y} \) for the input image \( \mathbf{x} \).
 Formally, let the vision-language model be denoted as \( f_{\text{LVLM}}\), and the inference process with \( k \) demonstrations of incomplete in-context learning is defined as follows:
\begin{equation}
\mathcal{D} = \{\mathbf{x}_i, y_i\}_{i=1}^k, \quad y_i \in \{C_1, C_2, \cdots, C_w\},
\end{equation}
\begin{equation}
\hat{y} = f_{\text{LVLM}}(\mathcal{D}, \mathbf{x}),
% \end{equation}
% \begin{equation}
\quad \hat{y} \in \{C_1, C_2, \cdots, C_m\},
\end{equation}
where \( \mathcal{D} = \{\mathbf{x}_i, y_i\}_{i=1}^k \) is the in-context data consisting of \( k \) demonstrations with input image \( \mathbf{x} \) and the corresponding label \( y_i \). These demonstrations, \( \{\mathbf{x}_i, y_i\} \), are retrieved from the incomplete retrieval database \( D_{\text{in}} \), so \( y_i \in \{C_1, C_2, \cdots, C_w\} \). Given a new image \( \mathbf{x} \), VICL instructs the LVLMs to generate a prediction \( \hat{y} \), which lie within the set \( \{C_1, C_2, \cdots, C_m\} \).

\end{definition}

\subsection{Empirical Study of IICL}\label{Empirical_Study}

\begin{figure}[ht]
    \centering
    \begin{subfigure}{0.32\linewidth}  
        \centering
        \includegraphics[width=1\linewidth,height=2cm]{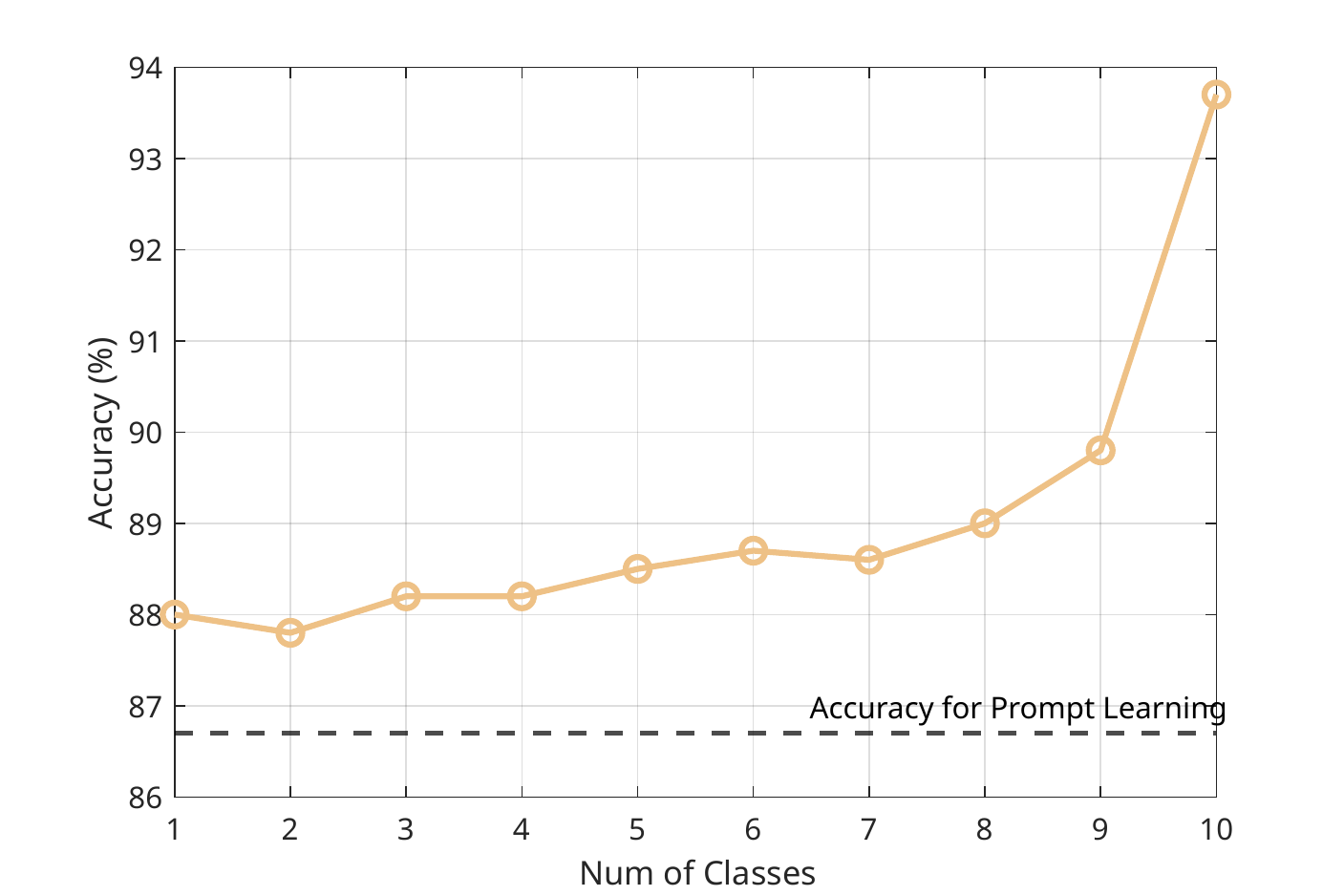}  
        \caption{}  
        \label{subfig_a}  
    \end{subfigure}
    \hfill  
    \begin{subfigure}{0.32\linewidth}  
        \centering
        \includegraphics[width=1\linewidth,height=2cm]{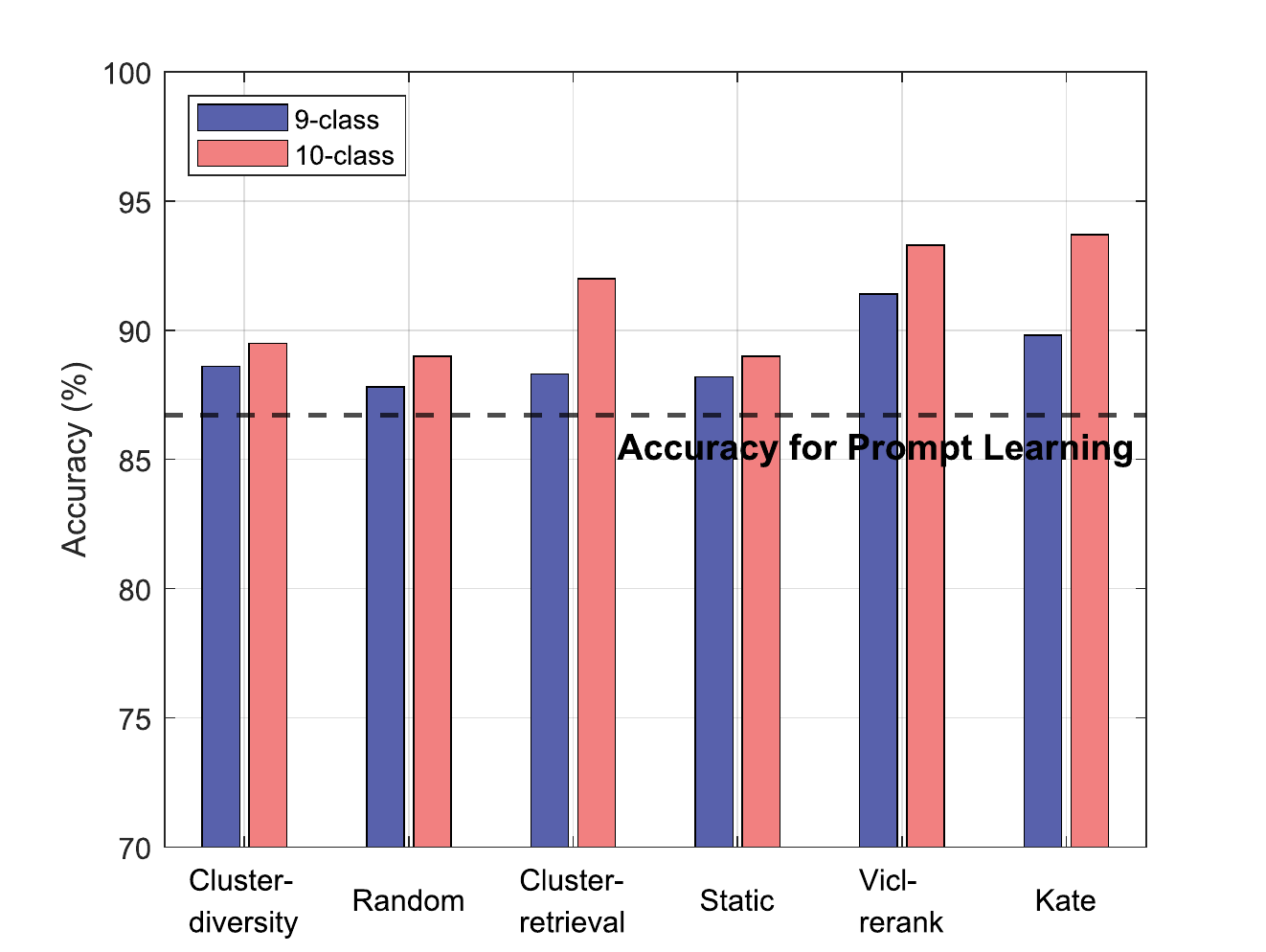}  
        \caption{}  
        \label{subfig_b}  
    \end{subfigure}
    \hfill
    \begin{subfigure}{0.32\linewidth}  
        \centering
        \includegraphics[width=1\linewidth,height=2cm]{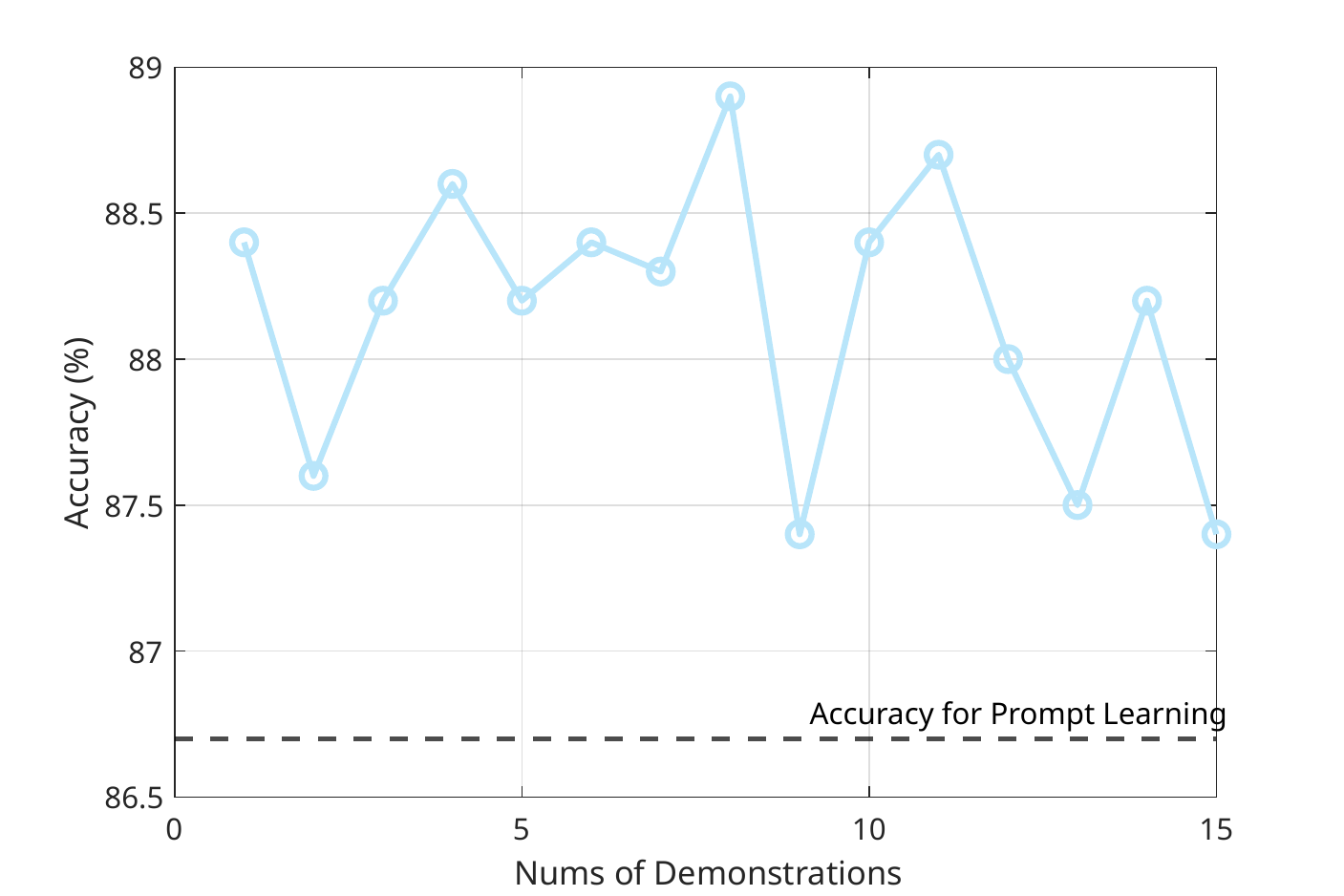}  
        \caption{}  
        \label{subfig_c}  
    \end{subfigure}
    \caption{Subfigure (a) shows the empirical study of IICL with different missing label numbers. Subfigure (b) shows the empirical study of IICL with different VICL methods. Subfigure (c) shows the empirical study of IICL with different demonstration numbers.}
    \label{fig_comparison}
\end{figure}

In this section, we examine the impact of an incomplete retrieval database and IICL for image classification. We conduct experiments using the CIFAR-10~\cite{abouelnaga2016cifar} dataset and the InternVL 2.5-8B~\cite{chen2024expanding,chen2024internvl} LVLMs. 

We utilize the CIFAR-10 training set as the retrieval database, which consists of data associated with 10 distinct labels. To simulate varying levels of incompleteness, we progressively reduce the number of available labels, decreasing from 10 labels to 9 labels, and ultimately to a single label.
When all 10 labels are present, the retrieval database encompasses a complete set of labeled data. Conversely, when only one label is available, data scarcity is at its most severe. Meanwhile, we employ several VICL methods, including Static, Random, Clustering-retrieval~\cite{li2023mot}, Kate~\cite{liu2022makes}, Cluster-Diversity~\cite{naik2023diversity}, and Vicl-rerank~\cite{zhou2024visual}, to investigate their performance under scenarios involving the absence of one label and 10 demonstrations.
Furthermore, we investigate the impact of the number of demonstrations on VICL performance under conditions of substantial data absence. Specifically, within a retrieval database containing only one label, we examine how increasing the number of demonstrations from 1 to 15 affects VICL performance.

\textbf{The performance of IICL declines as the number of missing labels increases, although it remains superior to that of the zero-shot prompt.} Figure~\ref{subfig_a} presents the empirical results of the IICL  applied to an incomplete retrieval database with Kate~\cite{liu2022makes}  method.
% 我们发现随着越多标签对应的数据开始缺失，IICL的表现逐渐变差，但是一直好于zero-shot prompt的表现。具体来说，哪怕incomplete retrieval database中只有一种标签对应的数据，IICL依然达到了xx\%的准确率，大于zero-shot promptxx\%的准确率。
We observe that as more label-corresponding data become missing, the performance of IICL gradually declines; however, it consistently outperforms the zero-shot prompt. Specifically, even when the incomplete retrieval database contains data for only a single label, IICL achieves an accuracy of 88.5\%, surpassing the zero-shot prompt, which attains 86.7\% accuracy. Furthermore, we observe that the removal of just one label results in a substantial accuracy drop, decreasing from 93.7\% to 88.5\%.

\textbf{All methods exhibit a decrease in accuracy when data corresponding to one label is missing, although their performance still remains superior to that of the zero-shot prompt baseline.} As illustrated in Figure~\ref{subfig_c}, the accuracy of all six methods declines under the one-label-missing scenario; however, they consistently achieve higher accuracy than the zero-shot prompt accuracy of 86.7%.

\textbf{IICL performs better with fewer demonstrations than with a larger number. However, regardless of the number of demonstrations, IICL consistently outperforms the zero-shot prompt.} 

As illustrated in Figure~\ref{subfig_c}, we conduct experiments under the most extreme label-missing scenario, where the incomplete retrieval database contains data corresponding to only a single label. By varying the number of demonstrations from 1 to 15, we observe that the accuracy of IICL stabilizes within a range of 87\% to 89\%. These results indicate that increasing the number of demonstrations does not mitigate the performance degradation associated with IICL.

Our empirical study demonstrates that an incomplete retrieval database significantly reduces IICL performance in image classification tasks. As the number of missing labels increases, IICL accuracy declines. More importantly, 
employing the different VICL methods and expanding the demonstration set do not effectively mitigate this issue. Nevertheless, regardless of the number of missing labels or demonstrations, IICL consistently outperforms the zero-shot prompt. This finding underscores the need for developing missing label-robust methods to enhance the performance of IICL against an incomplete retrieval database.

\section{Observation and Motivation}\label{Observation_Iterative_Judgment}
As the Figure \ref{subfig_b1} shows,
a three-class image classification problem, categorizing the image as cat, dog, or fish, can be reformulated as a sequence of three binary classification tasks: determining whether an image belongs to the ``Cat'', ``Dog'', or ``Fish''.

% \usepackage{booktabs}
\begin{table}[h]
\centering
\caption{The ground-truth label and other labels of the image.}
\label{other-label}
\resizebox{0.4\textwidth}{!}{%
\begin{tabular}{@{}c|c@{}}
\toprule
Ground-truth Label & Other Labels      \\ \midrule \midrule
Dog   & Not cat, Not fish \\
Cat   & Not dog, Not fish \\
Fish  & Not dog, Not cat  \\ \midrule
$C_j$  & \( \overline{C_1}, \overline{C_2}, \cdots, \overline{C_{j-1}}, \overline{C_{j+1}}, \cdots, \overline{C_m} \)             \\ \bottomrule
\end{tabular}
}
\end{table}

As illustrated in Table \ref{other-label}, an image labeled as ``Cat'' is also categorized as ``Not Dog'' and ``Not Fish''. In an incomplete retrieval database containing only “Cat'' and “Dog'' data, a standard three-class VICL lacks representative demonstrations for the ``Fish'' label. However, within the binary classification sequence, the ``Cat'' and “Dog'' tasks include both positive and negative demonstrations, and the binary classification task of ``Fish'' also contains ``Not Fish'' demonstrations, thereby ensuring a complete VICL process.
Therefore, leveraging a sequence of binary classification tasks facilitates the transformation of an incomplete retrieval database into a complete one and reformulates IICL into a fully realized VICL process.

During an exam, individuals typically complete the test first and then proceed to review their responses. In the review phase, they reflect on their answers based on both the questions and their initial responses. For example, when revisiting multiple-choice questions (where the correct answer is selected from options A, B, C, and D), individuals may reconsider those they initially found difficult. For questions that have already been answered, they might skip them to save time or reassess their answers using alternative methods to ensure consistency with their original choice. When uncertain about a question, individuals often narrow down the options and continue their reasoning within the reduced set. For instance, when faced with options A, B, C, and D, they might eliminate C and D but remain uncertain between A and B, continuing to deliberate between these two choices.

The example in Section \ref{Observation_Iterative_Judgment} is summarized as follows:  
An \( m \)-class image classification task with labels \( \{C_1, C_2, \cdots, C_m\} \) can be reformulated as a sequence of binary classification tasks, each determining whether an image belongs to \( C_1, C_2, \cdots, C_m \). If an image's true label is \( C_j \), it also is  not any other \( C_i \) (\( i \neq j \)), which can be represented as \( \overline{C_1}, \overline{C_2}, \cdots, \overline{C_{j-1}}, \overline{C_{j+1}}, \cdots, \overline{C_m} \).  
In the IICL scenario, the incomplete retrieval database \( D_{\text{in}} \) contains data for only \( w \) labels, where \( w < m \). For the \( j \)-th binary classification task (\( 1 \leq j \leq m \)), LVLMs employ VICL to determine whether the input image \( \mathbf{x} \) belongs to \( C_j \) or \( \overline{C_j} \), and the corresponding demonstrations' form changes from \([ \text{Image}, y, y \in \{C_1, C_2, \cdots, C_w\} ]\) to \([ \text{Image}, y^{j}, y^{j} \in \{C_j, \overline{C_j}\} ]\). If \( j \leq w \), the database contains data for both \( C_j \) and \( \overline{C_j} \); otherwise, it contains at least \( \overline{C_j} \). Demonstrations from \( \overline{C_j} \) further inform LVLMs about the exclusion of \( C_j \).  The process of LVLMs employ VICL to predict the true label \( y \) is as follows:
\begin{equation}
\begin{array}{c}
\mathcal{D}_{\text{IJ}}^j = \{\mathbf{x}_i^d, y_i^{j}\}_{i=1}^k, \quad 0 \leq j \leq m, \quad y_i^j \in \{C_j, \overline{C_j}\}, \\ \mathcal{D}_{\text{IJ}}=\{\mathcal{D}_{\text{IJ}}^1, \mathcal{D}_{\text{IJ}}^2, \cdots, \mathcal{D}_{\text{IJ}}^m \}
\end{array}
\end{equation}
\begin{equation}
\begin{array}{c}
\hat{y}^1_{\mathbf{x}}, \hat{y}^2_{\mathbf{x}}, \cdots, \hat{y}^m_{\mathbf{x}} = f_{\text{LVLM}}^2(\mathcal{D}_{\text{IJ}}, \mathbf{x}),
 \hat{y}^j \in \{C_j, \overline{C_j}\}, \\
 % \mathcal{\hat{y}}=\{[\hat{y}^1, \hat{y}^2, \cdots, \hat{y}^m]\}
\end{array}
\end{equation}\label{two_class_equa}

where $f_{\text{LVLM}}^2$ is the sequence of binary classification tasks.
\( \mathcal{D}_{\text{IJ}}^j \) denotes the \( k \) demonstrations for the \( j \)-th sub-question. 

We assume that in  \( m \)-class classification, LVLM can employ VICL to predict the true label \( y \) of input image \( \mathbf{x} \) through a sequence of  $m$ binary classification tasks, which is presented in Assumption \ref{assumption:ova-ovo-LVLM}.

\begin{assumption}
\label{assumption:ova-ovo-LVLM}
LVLMs can approximate the true label \( y_{\mathbf{x}} \) of $\mathbf{x}$ through $m$ binary classifications. Let
\( \mathbf{x} \) be the input image, \( f_{\text{LVLM}} \) denote the function of LVLMs for the binary classifications,  \(\hat{y}^1_{\mathbf{x}}, \hat{y}^2_{\mathbf{x}}, \cdots, \hat{y}^m_{\mathbf{x}} \) are the predict labels of $m$ binary classifications, and we assume that the predicted label \( \hat{y}_{\mathbf{x}} \) approximates the true label \( y_{\mathbf{x}} \). 
% where \( \mathbf{x} \) represents the input image, \( f_{\text{LVLM}}^2 \) denotes the Vision-Language Model (LVLM) for the binary classification task, and \( \hat{y}^j_{\mathbf{x}} \) is the predicted binary label for image \( \mathbf{x} \) in the \( j \)-th task. Subsequently, we predict the true label of \( \mathbf{x} \) based on the predictions from the series of tasks, \( \hat{y}^1_{\mathbf{x}}, \hat{y}^2_{\mathbf{x}}, \cdots, \hat{y}^m_{\mathbf{x}} \)
Formally:
\begin{equation}
\hat{y}_{\mathbf{x}}=f_{\text{LVLM}}( \mathbf{x}, \hat{y}^1_{\mathbf{x}}, \hat{y}^2_{\mathbf{x}}, \cdots, \hat{y}^m_{\mathbf{x}}), \| \hat{y}_{\mathbf{x}}- y_{\mathbf{x}}\|_p \leq \epsilon,
\end{equation}
where \(\epsilon\) is a sufficiently small constant. The condition 
$
\|\hat{y}_{\mathbf{x}} - y_{\mathbf{x}}\|_p \leq \epsilon
$
implies that the difference between \(\hat{y}_{\mathbf{x}}\) and \(y_{\mathbf{x}}\) is negligible. Consequently, the predicted label \(\hat{y}_{\mathbf{x}}\) serves as a close approximation of the true label \(y_{\mathbf{x}}\). The notation \(\|\cdot\|_p\) denotes a distance metric, such as the \(l_2\)-norm.
\end{assumption}
% \section{Iterative Judgment and Integrated Prediction}
\section{Method}

To mitigate the change caused by incomplete retrieval database and IICL, we propose the \textit{Iterative Judgments and Integrated Prediction (IJIP)} method, which transforms incomplete retrieval database and IICL into a complete retrieval database and VICL. IJIP is a two-stage in-context learning method:
\ding{182} \textbf{Iterative Judgment Stage} – Reformulates an \( m \)-classification task into \( m \) binary classification sub-questions, where \( j \)-th sub-question prompts the LVLMs to determine whether the label of the given image is  \( C_j \) or not. 
\ding{183} \textbf{Integrated Prediction Stage} – Combines the image with its Iterative Judgment predictions and applies \( m \)-classification VICL to finalize the label.

\subsection{Iterative Judgment Stage}

In the Iterative Judgment Stage, drawing inspiration from methods like KATE~\cite{liu2022makes} and VICL-rerank~\cite{zhou2024visual}, we retrieve \( k \) labeled images from the incomplete retrieval database that are most semantically similar to the input image and use them as demonstrations. The retrieval process comprises the following steps:
\ding{182}
\textit{Vectorization}: The input image $\mathbf{x}$ and all labeled data $\mathcal{D} = \{(\mathbf{x}_1, y_1), (\mathbf{x}_2, y_2), \cdots, (\mathbf{x}_n, y_n)\}$ in the incomplete retrieval database are encoded into vector representations using the pre-trained CLIP model $f_{\text{pre}}$.
\ding{183}
\textit{Similarity Computation}: Cosine similarity is computed between the input image and all labeled data.
\ding{184}
\textit{Demonstration Selection}: The \( k \) most similar labeled images are chosen as demonstrations. The image with the highest similarity is assigned as the first demonstration, while the image with the \( w \)-th highest similarity is placed in the \( w \)-th position. Formally:
 \begin{equation}
 \mathbf{e}_\mathbf{x}=f_{\text{pre}}(\mathbf{x}), \mathbf{e}_i=f_{\text{pre}}(\mathbf{x}_i),
\mathbf{E} = \{\mathbf{e}_1, \mathbf{e}_2, \cdots, \mathbf{e}_n\}. 
\end{equation}
\begin{equation}
s_i=\frac{\mathbf{e}_\mathbf{x} \cdot \mathbf{e}_i}{\|\mathbf{e}_\mathbf{x}\| \cdot \|\mathbf{e}_i\|},
% =
% \frac{f_{\text{pre}}(\mathbf{x}) \cdot f_{\text{pre}}(\mathbf{x}_i)}
% {\|f_{\text{pre}}(\mathbf{x})\| \cdot \|f_{\text{pre}}(\mathbf{x}_i)\|},
\mathcal{S} = \{s_1, s_2, \cdots, s_n\}. 
\end{equation}
% \begin{equation}
% \small
%     \mathcal{I} = \arg\operatorname{top}^k_{(\mathbf{x}_i, \mathbf{y}_i)\subseteq\mathcal{D}} \mathcal{S}, \mathcal{D}_d = \{\mathbf{x}_i^d, y_i^{d}\}_{i=1}^k, \text{for }i \in \mathcal{I},
% \end{equation}

\begin{equation}\label{D_d}
\small
\begin{array}{c}
\mathcal{I} = \arg\operatorname{top}^k_{(\mathbf{x}_i, \mathbf{y}_i)\subseteq\mathcal{D}} \mathcal{S}, \\
 \mathcal{D}_d = \{\mathbf{x}_i^d, y_i^{d}\}_{i=1}^k, y_i^{d} \in \{C_1, C_2, \cdots, C_w\}, \text{for }i \in \mathcal{I},
\end{array}
\end{equation}
where $\mathbf{e}_\mathbf{x}$ is the embedding of  $\mathbf{x}$, and $\mathbf{e}_i$
is the embedding of  $\mathbf{x}_i$, and $\mathbf{x}_i$ is an image of the incomplete retrieval database. $s_i$ is the cosine similarity of input image $\mathbf{x}$ and image $\mathbf{x}_i$. $\mathcal{I}$ is the index list of the \( k \) most similar labeled images, and  $\mathcal{D}_d$ is the demonstrations of $\mathbf{x}$. $\{C_1, C_2, \cdots, C_w\}$ is the accessible labels in the incomplete retrieval database.

As illustrated in Figure \ref{overview}, the \( m \)-class classification problem is transformed into a sequence of binary classification tasks, with each task corresponding to one of the \( m \) sub-classification problems. In the \( j \)-th sub-classification task, the goal is to determine whether the label of the given image is \( C_j \) or \( \overline{C_j} \). Consequently, the demonstration format is modified as follows:
$
\mathcal{D}_{\text{IJ}}^j = \{\mathbf{x}_i^d, y_i^{j}\}_{i=1}^k,
$
where \( 0 \leq j \leq m \) and \( y_i^j \in \{C_j, \overline{C_j}\} \). 

Then, based on Equation \ref{two_class_equa}, the binary classification results \( \hat{y}^1, \hat{y}^2, \dots, \hat{y}^m \) are obtained using the sequential binary classifier \( f_{\text{LVLM}}^2 \).

\begin{tcolorbox}
\textbf{Remark:}

In the Iterative Judgment Stage, we reformulate the 
$m$-class classification task into a series of binary classification tasks. Although this reformulation comprises 
$m$ sub-questions, the LVLMs are queried only once during this stage. Specifically, all 
$m$ sub-questions and their corresponding retrieved demonstrations are integrated into a single consolidated prompt, enabling simultaneous responses from the LVLMs.
\end{tcolorbox}

\subsubsection{Analysis of Iterative Judgments Stage}
% \subsubsection{Analysis of iterative judgments stage.}
In this section, we analyze the role of the iterative judgments stage in refining the classification process.

\paragraph{Transforming Incomplete In-context Learning into In-context Learning.}
During the iterative judgments stage, we reformulate the \( m \)-class classification problem as a sequence of binary classification tasks. In each \( j \)-th binary classification task, the goal is to determine whether the given image belongs to category \( C_j \). Since the retrieval database always contains at least one instance of \( \overline{C_j} \), none of these \( m \) binary classification tasks encounter missing categories. Thus, the iterative judgments stage effectively converts Incomplete In-context Learning into standard In-context Learning.

\paragraph{Reduction in Classification Complexity: \( m \) Binary Classification Tasks vs. a Single \( m \)-Class Classification Task.}  
By decomposing the classification problem into \textit{m} binary classification tasks, each \( j \)-th task requires a binary decision—whether the given image belongs to category \( C_j \) or not. This approach is inherently less complex than directly identifying the correct label among \( m \) possible classes in a conventional \( m \)-class classification task. For example, determining whether an image contains a cat (binary classification) is inherently simpler than identifying the exact species of the depicted animal (multi-class classification).

\subsection{Integrated Prediction Stage}

We refine the classification decision based on the given image $\mathbf{x}$ and the corresponding predictions \( \{\hat{y}^1_{\mathbf{x}}, \hat{y}^2_{\mathbf{x}}, \cdots, \hat{y}^m_{\mathbf{x}} \}\) from the Iterative Judgment Stage. Specifically, we first determine the number of predictions  among \( \{\hat{y}^1_{\mathbf{x}}, \hat{y}^2_{\mathbf{x}}, \cdots, \hat{y}^m_{\mathbf{x}} \}\) that correspond to the labels \( \{C_1, C_2, \cdots, C_m\} \). The computation is defined as follows:
\begin{equation}
\mathbb{I}_{\mathbf{x}}^{j} =
\begin{cases}
1, & \text{if } \hat{y}^j = C_j, \\
0, & \text{otherwise}
\end{cases}
\quad \mathbb{I}_{\mathbf{x}}=\sum_{i=1}^{m} \mathbb{I}_{\mathbf{x}}^{j},
\end{equation}
where $\mathbb{I}_{\mathbf{x}}^{j}$ is an indicator function, $\mathbb{I}_{\mathbf{x}}$ is the total number of $\hat{y}^j = C_j$.

As illustrated in Figure \ref{overview}, when \( \mathbb{I}_{\mathbf{x}} = 0 \), we apply the IICL of the \( m \)-class classification $f_{\text{LVLM}}^m$ and the demonstrations $\mathcal{D}_d$  to the input image \( \mathbf{x} \) to obtain its predicted label.
When \( \mathbb{I}_{\mathbf{x}} = 1 \), we assume that prediction from the \( j \)-th binary classification task identifying \( C_j \), while the remaining \( m-1 \) binary classification tasks classify \( \mathbf{x} \) as \( \overline{C_1}, \overline{C_2}, \cdots, \overline{C_{j-1}}, \overline{C_{j+1}}, \cdots, \overline{C_m} \). In this case, \( C_j \) is directly assigned as the predicted label for \( \mathbf{x} \).
If \( \mathbb{I}_{\mathbf{x}} > 1 \), it indicates that \( \mathbf{x} \) belongs to multiple labels according to different binary classification tasks. Suppose \( \mathbb{I}_{\mathbf{x}} = u \); 
in the Integrated Prediction Stage, the LVLM employs the IICL of the \( u \)-class classification $f_{\text{LVLM}}^u$ and the demonstrations $\mathcal{D}_d$, whose candidate labels are the corresponding labels for $\mathbb{I}_{\mathbf{x}}^{j} =1$, to the input image \( \mathbf{x} \). 
 For example, if the predicted labels for \( \mathbf{x} \) across the \( m \) binary classification tasks include \( C_i, C_j \), and \( \overline{C_1}, \overline{C_2}, \cdots, \overline{C_{i-1}}, \overline{C_{i+1}}, \cdots, \overline{C_{j-1}}, \overline{C_{j+1}}, \cdots, \overline{C_m} \), then the LVLMs select the candidate labels for \( \mathbf{x} \) from \( C_i \) and \( C_j \).  

 Formally: 
\begin{equation}\label{2-stage}
\hat{y}_{\mathbf{x}}= 
\begin{cases}
f_{\text{LVLM}}^m( \mathcal{D}_d,\mathbf{x}), & \text{if } \mathbb{I}_{\mathbf{x}} = 0, \\
C_j, \textit{s.t.} \mathbb{I}_{\mathbf{x}}^j = 1  & \text{if } \mathbb{I}_{\mathbf{x}} = 1, \\
f_{\text{LVLM}}^u(\mathcal{D}_d, \mathbf{x}),    & \text{if } \mathbb{I}_{\mathbf{x}} = u, 1<u\leq m,\\
\end{cases}
\end{equation}
where $\mathcal{D}_d$ is the demonstrations from the Equation \ref{D_d},
$\mathcal{D}_d = \{\mathbf{x}_i^d, y_i^{d}\}_{i=1}^k, y_i^{d} \in \{C_1, C_2, \cdots, C_m\}$
When \( \mathbb{I}_{\mathbf{x}} = 1 \), we assign the corresponding label \( C_j \), where \( \mathbb{I}_{\mathbf{x}}^j = 1 \), as the final predicted label for \( \mathbf{x} \).
% 其中,f_{\text{LVLM}}^k表示在k个
\subsubsection{Analysis of Integrated Prediction Stage}

In this section, we evaluate the effectiveness of the strategies associated with the three cases in Equation \ref{2-stage}.

\textbf{A new prediction opportunity for  $\mathbb{I}_{\mathbf{x}} = 0$ scenario.}

When \(\mathbb{I}_{\mathbf{x}}^j = 0\), the Iterative Judgments Stage is unable to provide a definitive prediction. In this case, we apply IICL using the corresponding demonstration set \(\mathcal{D}_d\) to perform an \( m \)-class classification. This approach effectively allows the image \( \mathbf{x} \) an additional opportunity for prediction.

\textbf{The result of multiple confirmations for  $\mathbb{I}_{\mathbf{x}} = 1$ scenario.}

When \( \mathbb{I}_{\mathbf{x}}^j = 1 \), we assume \( \mathbb{I}_{\mathbf{x}}^{j} =1 \), indicating that the Iterative Judgments Stage predicts \( C_j \) while also confirming  
\( \overline{C_1}, \overline{C_2}, \dots, \overline{C_{j-1}}, \overline{C_{j+1}}, \dots, \overline{C_m} \).  
Compared to direct \( m \)-class classification, IJIP not only identifies the given image as belonging to \( C_j \) but also explicitly verifies that it does not belong to \( C_1, C_2, \dots, C_{j-1}, C_{j+1}, \dots, C_m \).  
Thus, the classification decision in IJIP is strengthened through multiple confirmations, enhancing its prediction success probability compared to direct \( m \)-class classification.

% \paragraph{Eliminate the least probable options to narrow the selection range for  $\mathbb{I}_{\mathbf{x}} = u$ scenario.}
\textbf{Eliminate the least probable options to narrow the selection range for  $\mathbb{I}_{\mathbf{x}} = u$ scenario.}

When \( \mathbb{I}_{\mathbf{x}} = u \), if \( u = m \), then \( f_{\text{LVLM}}^u(\mathcal{D}_d, \mathbf{x}) \) corresponds to performing an \( m \)-class IICL task for \( \mathbf{x} \). However, when \( u < m \), the Iterative Judgment Stage eliminates less probable options, thereby reducing the selection space.
This staged prediction process mirrors human cognitive decision-making, which progresses from broad elimination to fine-grained reasoning. For instance, in multiple-choice exams, test-takers typically begin by discarding unlikely options before carefully evaluating the remaining choices. Specifically, when presented with four options (A, B, C, and D), a student might initially eliminate C and D through the elimination process and then refine their choice by selecting between A and B.

\section{Experiments}
\subsection{Experimental Setup}
\textbf{Datasets.}
We conduct experiments using the CIFAR-10~\cite{krizhevsky2009learning} and Fashion-MNIST~\cite{xiao2017fashion} datasets.
Detailed information about the datasets is provided in Section \ref{datasets} of the Appendix.

\noindent\textbf{Metrics.} 
We employ accuracy as the metric, with higher accuracy reflecting greater performance.
Detailed information about the datasets is provided in Section \ref{metric} of the Appendix.

\noindent \textbf{Baselines.}
We perform experiments using various ICL and VICL methods, including  Static, Random, Clustering-retrieval~\cite{li2023mot}, Kate~\cite{liu2022makes}, Cluster-Diversity~\cite{naik2023diversity}, and Vicl-rerank~\cite{zhou2024visual}. Detailed information about the datasets is provided in Section \ref{BASE} of the Appendix.

\noindent \textbf{Vision-language models and other setup.}
The Vision-language model chosen for our experiments is InternVL 2.5-8B~\cite{chen2024expanding}  and InternVL 2.5-4B~\cite{chen2024expanding}.
The pre-trained model employed in this study is CLIP (ViT-Base-Patch32)~\cite{radford2021learning}. Each experiment is conducted three times, and the average result is reported.

\subsection{Main Results}

IJIP demonstrates strong performance across various experimental conditions. As shown in Table~\ref{missing_label_numbers}, even under the extreme scenario where data corresponding to only a single label is available, IJIP achieves an accuracy of 88.5\% on the CIFAR-10 dataset. Except for the scenario involving 90\% missing labels on Fashion-MNIST—where available data is severely limited—IJIP consistently outperforms all baseline methods across all experimental settings.

\begin{table}[t]
\centering
\caption{Comparsion IJIP with other VICL methods. We evaluate performance primarily using accuracy(\%)↑ under varying proportions of missing labels. The experiments are conducted with label missingness levels of \(10\%\), \(40\%\), and \(90\%\). \textbf{Each experiment is conducted three times, and the average result is reported.}
} 

\label{Main_Results_Table}
\resizebox{0.45\textwidth}{!}{%
\begin{tabular}{@{}c|c|ccc|ccc@{}}
\toprule
\multirow{2}{*}{Data}          & \multirow{2}{*}{Methods} & \multicolumn{3}{c|}{InternVL 2.5-8B}          & \multicolumn{3}{c}{InternVL 2.5-4B}           \\ \cmidrule(l){3-8} 
                               &                          & 90\%          & 40\%          & 10\%          & 90\%          & 40\%          & 10\%          \\ \midrule \midrule
\multirow{7}{*}{CIFAR-10 }       & Cluster-diversity        & 86.8          & 88.7          & 88.3          & 78.7          & 80.4          & 83.2          \\
                               & Random                   & 87.4          & 88.9          & 88.6          & 79.8          & 81.7          & 83.7          \\
                               & Cluster-retrieval        & 88.0          & 88.7          & 89.8          & 75.6          & 85.4          & 87.7          \\
                               & Static                   & 86.7          & 87.8          & 87.8          & 76.3          & 82.2          & 84.3          \\
                               & Vicl-rerank              & 88.1          & 88.9          & 88.2          & 74.0            & 80.0          & 90.8          \\
                               & Kate                     &88.0          & 88.7          & 89.8          & 75.0            & 80.3          & 90.1          \\ \cmidrule(l){2-8} 
                               & \textbf{IJIP}     \cellcolor{lightpurple}                & \textbf{89.2}\cellcolor{lightpurple}& \textbf{91.5}\cellcolor{lightpurple}& \textbf{92.3}\cellcolor{lightpurple}& \textbf{88.5}\cellcolor{lightpurple}& \textbf{88.8}\cellcolor{lightpurple}& \textbf{93.9}\cellcolor{lightpurple}\\ \midrule
\multirow{7}{*}{Fashion-MNIST} & Cluster-diversity        & 52.4             & 53.8          & 54.0           & 52.6          & 54.2          & 52.9          \\
                               & Random                   & 52.2          & 55.7          & 54.5          & 51.8          & 54.1          & 53.3          \\
                               & Cluster-retrieval        & 46.6          & 56.6          & 57.8          & 46.7          & 51.6          & 49.6          \\
                               & Static                   & 49.9          & 53.1          & 54.6          & 51.4          & 52.8          & 50.3          \\
                               & Vicl-rerank              & 43.2             & 60.9          & 75.7          & 41.1          & 61.3          & 72.4          \\
                               & Kate                     & 42.3          & 60.7          & 74.6          & 42.9          & 63.3          & 73.6          \\ \cmidrule(l){2-8} 
                               & \textbf{IJIP}\cellcolor{lightpurple}& \textbf{52.8}\cellcolor{lightpurple}& \textbf{68.9}\cellcolor{lightpurple}& \textbf{78.9}\cellcolor{lightpurple}& 47.5\cellcolor{lightpurple}& \textbf{64.1}\cellcolor{lightpurple}& \textbf{77.4}\cellcolor{lightpurple}\\ \bottomrule
\end{tabular}
}
\end{table}

\subsection{Analysis}

We analyze the effects of demonstration number, missing-label proportion, and LVLM size on IJIP performance. Specifically, we vary the number of demonstrations from 1 to 10 with 10\% label missingness. To evaluate the impact of missing-label proportion, we fix the demonstration count at 10 and progressively increase missing labels in the incomplete retrieval database from 10\% to 90\%. Lastly, we assess the influence of LVLM size by employing InternVL 2.5-2B~\cite{chen2024internvl} and InternVL 2.5-26B~\cite{chen2024expanding} under consistent settings (10 demonstrations and 10\% missing labels).

\textbf{Different demonstration numbers.}

We investigate the relationship between the number of demonstrations and the performance of IJIP. As illustrated in Sub-figure~\ref{subfig_c}, accuracy improves with an increasing number of demonstrations, particularly for challenging datasets such as Fashion-MNIST when using less capable models (e.g., InternVL 2.5-4B). Specifically, the accuracy of InternVL 2.5-4B significantly increases from 55.2\% to 77.6\% as the number of demonstrations rises from 1 to 10.

\begin{figure}[!ht]
  \centering
   \includegraphics[width=0.75\linewidth]{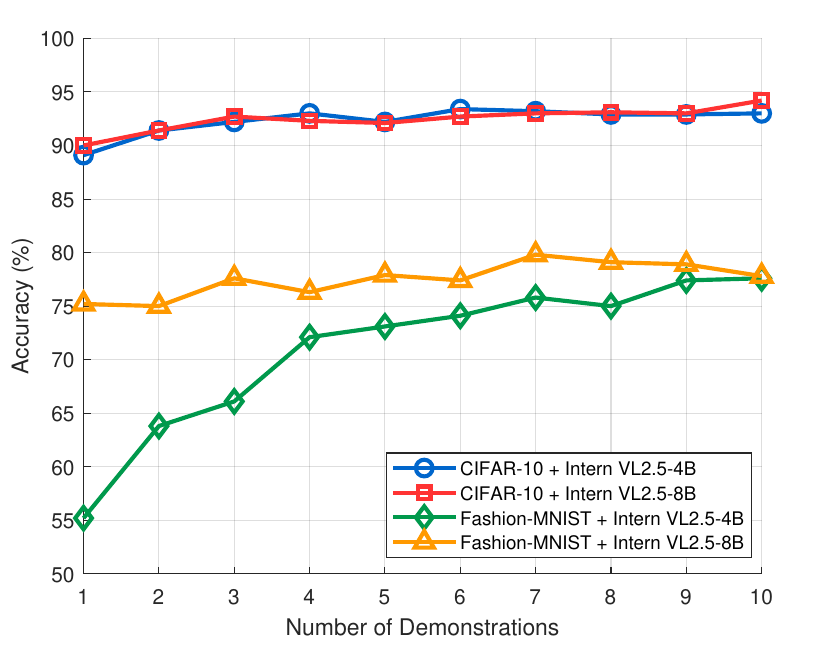}
   \caption{The accuracies(\%)↑ of different demonstration numbers. Each experiment is conducted three times, and the average result is reported.}
   \label{demonstration_numbers}
\end{figure}

\textbf{Different missing label proportions.}

We investigate the relationship between the performance of IJIP and the proportion of missing labels. The experimental results are presented in Table \ref{missing_label_numbers}. The accuracy of IJIP decreases as the proportion of missing labels increases, dropping from 
77.4\% to 
47.5\% on the Fashion-MNIST dataset using the InternVL 2.5-4B model.

\textbf{Different  LVLMs' sizes.}

We utilize four Intern VL2.5 models (2B, 4B, 8B, and 26B), where model size serves as an indicator of capability. As shown in Table~\ref{LVLMs_size} (Appendix), average accuracy consistently increases from 88.4\% to 97.3\% with larger model sizes.

\section{Discussion}
\subsection{Expanding IJIP from IICL to Complete VICL}

Previously, our method was applied to IICL. We now extend it to complete VICL, where the retrieval database contains data for all labels. Results in Table \ref{LVLMs_size} (Appendix) demonstrate that our method remains effective, achieving strong performance under complete VICL with an accuracy of up to 94.6\%.

% \usepackage{booktabs}
% \usepackage{multirow}

\subsection{Analyzing the Impact on IJIP from the Perspective of Classification Datasets}

The impact of a classification dataset on IICL primarily depends on task difficulty and label number. We assess difficulty via zero-shot prompt learning accuracy; lower accuracy indicates greater challenge. Besides CIFAR-10 and Fashion-MNIST, we analyze Wiki-Art~\cite{saleh2015large} Genre (24 classes) and Wiki-Art Artist~\cite{saleh2015large}(11 classes) datasets using IJIP. As shown in Table \ref{impact_dif_num}, Fashion-MNIST, despite having the same number of labels as CIFAR-10, is more challenging with a lower accuracy of 78.9\%. Conversely, Wiki-Art Genre, though having more categories than Wiki-Art Artist, achieves higher accuracy (73.8\%), indicating lower difficulty. Thus, lower task difficulty and fewer categories generally enhance classification accuracy, and IJIP effectively handles tasks even with numerous categories.
\begin{table}[h]
\centering
\caption{The impact of classification datasets. Each experiment is conducted three times, and the average result is reported.}
\label{impact_dif_num}
\resizebox{0.3\textwidth}{!}{%
\begin{tabular}{@{}c|ccc@{}}
\toprule
Dataset         & Zero-shot & IJIP & Labels \\ \midrule \midrule
CIFAR-10         & 86.7      & 92.3 & 10     \\
Fashion-MNIST   & 37.2      & 78.9 & 10     \\
Wiki-art genre  & 47.9      & 73.8 & 24     \\
Wiki-art artist & 33.4      & 65.2 & 11     \\ \bottomrule
\end{tabular}
}
\end{table}
\subsection{Expanding IJIP from ICL to Prompt Learning}

\begin{figure}[t]
\centering
\includegraphics[width=0.72\linewidth]{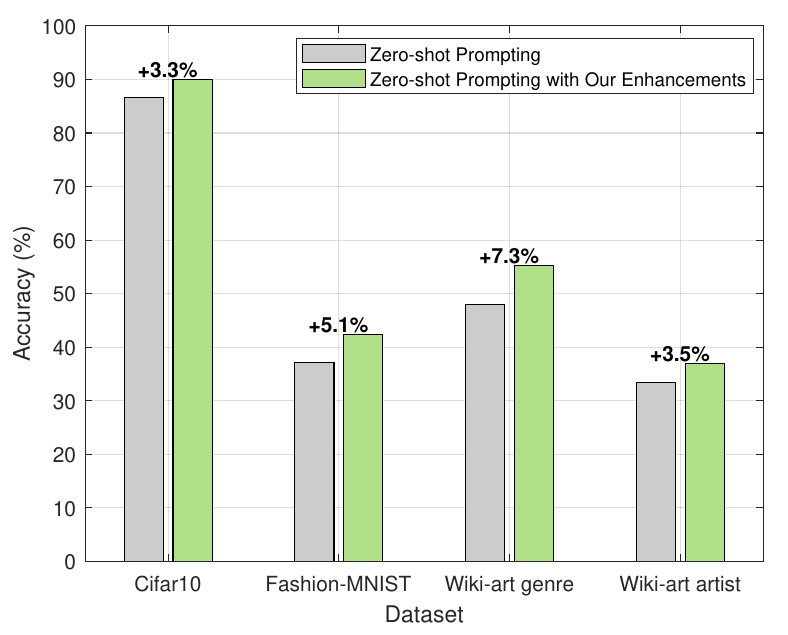}
\caption{The effectiveness of IJIP in improving zero-shot prompt learning performance. Each experiment is conducted three times, and the average result is reported.}

\label{zero-shot-prompt-results}
\end{figure}

In addition to context-based learning, prompt learning is a widely used method that leverages LVLMs. We integrate our method into prompt learning. The experimental results are provided in Figure \ref{zero-shot-prompt-results}. Our findings reveal that applying our method enhances the performance of prompt learning. Specifically, prompt learning experiences an average improvement of 4.8\%.

\subsection{Expanding IICL to Textual Modality}

In prior research, we primarily focused on Incomplete In-context Learning within the image modality. This work extends our approach to the text modality, using the SST5 and Emotion datasets. We conduct experiments on the Qiwen2.5-7B. Given that vicl-rerank~\cite{zhou2024visual} is restricted to the image domain, we use Zero-shot Prompt, Static, Random, Clustering Retrieval, Kate~\cite{liu2022makes}, and Cluster-Diversity as baselines. The results, presented in Table \ref{Textual_modality_table} (Appendix), show that our method achieves SOTA performance on both the SST5 and Emotion datasets, with accuracies of 62.3\% and 45.7\%, respectively. These findings demonstrate the transferability of our method to the text modality.
% \subsection{Expanding IICL from Classification to Input-to-Score Tasks}

% \usepackage{booktabs}

\section{Conclusion}

We introduce the concepts of the incomplete retrieval database and Incomplete In-context Learning (IICL) and conduct an empirical study to evaluate their impact. Additionally, we propose IJIP as a corresponding solution and examine the key factors contributing to its effectiveness. IJIP demonstrates strong performance across varying levels of label missingness, achieving notable results with 10\%, 40\%, and 90\% of labels absent from the retrieval database.
Moreover, even when all labels are available, IJIP outperforms all six baseline models. Furthermore, IJIP can be seamlessly integrated into Prompt Learning and adapted to the text domain, highlighting its versatility.
{
    \small
    \bibliographystyle{ieeenat_fullname}
    \bibliography{main}
}

\clearpage
\setcounter{page}{1}
\maketitlesupplementary
\appendix

\section{Related Work}

\paragraph{In-context learning and visual In-context learning.}
% \paragraph{In-Context Learning and Visual In-context Learning.}
In-context learning (ICL) is an innovative paradigm in which large language models (LLMs) make predictions based on context and a few demonstrations \citep{BrownICL, mavromatis2023examples, milios-etal-2023-context, min-etal-2022-metaicl}. However, ICL performance can be unstable, varying from state-of-the-art results to unpredictable outputs, depending on factors such as the selected demonstrations \citep{cheng2024exploring, fei-etal-2023-mitigating, gupta2023robust, li-qiu-2023-finding, lyu-etal-2023-z, min-etal-2022-rethinking, wei2023larger, wei2023symbol}, the order of demonstrations \citep{lu-etal-2022-fantastically, wu-etal-2023-self}, the templates used for demonstrations \citep{min-etal-2022-rethinking}, and the labels assigned to the demonstrations \citep{wang-etal-2023-label}.
Similarly, visual In-context Learning (VICL) is a paradigm comprising three core components: Visual Demonstration Retrieval, Intent-Oriented Image Summarization, and Intent-Oriented Demonstration Composition. VICL builds upon the concept of in-context learning, wherein large language models (LLMs) generate accurate responses by simply adjusting the input prompt, without the need for fine-tuning model parameters ~\cite{gpt1,gpt2,t5}. This emergent capability has been shown to be effective when scaled ~\cite{Wei2022EmergentAO,Wei2022ChainOT,Fu2022ComplexityBasedPF}. In the context of visual-language models, VICL adapts the in-context learning framework for visual question-answering tasks by creating input prompts that integrate visual demonstrations, query images, and text-based queries. The Visual Demonstration Retrieval (VDR) component of VICL seeks to identify relevant visual demonstrations by leveraging both the visual features of images and their textual descriptions. The retrieval process employs a pre-trained image encoder (e.g., ViT ~\cite{DosovitskiyB0WZ21}) to map images into a high-dimensional feature space and retrieve the top-n most relevant images. Cross-modal reranking then refines the selection, ensuring semantic relevance through a vision-language model such as CLIP ~\cite{RadfordKHRGASAM21}, which compares textual descriptions of the query and candidate images. This integrated visual-textual approach ensures that the retrieved demonstrations are both visually and contextually aligned with the task at hand ~\cite{Zhou0GTXLJJ23}.

\textbf{Class-Incremental Learning} (CIL)~\cite{zhang2025few,cao2025class,liu2025class,li2025prompt} enables models to learn new categories from sequential data while retaining knowledge of previous ones. The model initially learns from a subset of classes and adapts as new classes are introduced, without forgetting the earlier ones. CIL represents a scenario where the dataset does not include all potential categories during training.
\textbf{Extreme Class-Imbalance Learning} occurs when class distributions are highly imbalanced, with the minority class containing very few or no samples~\cite{attenberg2010label, akhbardeh2021handling}. This situation is more severe than typical class imbalance problems, where the minority class is underrepresented but still present.

In summary, prior research on ICL and VICL has primarily focused on the selection of demonstrations and their integration into an appropriate context. However, these studies have largely overlooked potential anomalies in the context retrieval database. For example, in Class-Incremental Learning scenarios, the context retrieval database may lack data for newly introduced classes. Additionally, in cases of extreme class imbalance, certain labels may not have corresponding datasets. Furthermore, due to data labeling issues, some labels may lack annotated data altogether. These real-world challenges underscore the importance of exploring context learning in scenarios where data corresponding to specific labels is missing.

\section{Datasets}\label{datasets}
\paragraph{CIFAR-10 dataset.}
The CIFAR-10 dataset is a widely used small-scale image collection comprising 60,000 32x32 color images distributed across 10 categories, with 6,000 images per category. These images are divided into 50,000 training samples and 10,000 test samples. The dataset presents a range of challenges, including variations in angles, poses, lighting, and backgrounds, making it difficult for both machine learning and deep learning algorithms. Due to its manageable size and pixel value normalization during preprocessing, CIFAR-10 is commonly used for benchmarking image classification tasks in computer vision.

\paragraph{Fashion-MNIST dataset.}
The Fashion-MNIST dataset is an alternative to the MNIST dataset for handwritten digit recognition, consisting of a collection of 60,000 training samples and 10,000 test samples. Each sample is a 28x28 grayscale image, similar in size to those in the MNIST dataset, but instead of digits, they represent images of fashion items from 10 categories: T-shirt/top, Trouser, Pullover, Dress, Coat, Sandal, Shirt, Sneaker, Bag, and Ankle boot. Fashion-MNIST was developed by employees of Zalando, a German online fashion retailer, and is intended for machine learning research. Fashion-MNIST is often used as a benchmark dataset in machine learning projects, particularly in introductory tutorials on deep learning and image recognition, due to its moderate complexity. The dataset can be easily accessed and downloaded through machine learning libraries like TensorFlow and PyTorch.

\begin{table}[h]
    \centering
    \small
    \begin{tabular}{c>{\centering\arraybackslash}p{2.2cm} >{\centering\arraybackslash}p{1.2cm} >{\centering\arraybackslash}p{2.2cm}} 
        \toprule
        \multicolumn{1}{c}{\textbf{Dataset}} & \multicolumn{1}{c}{\textbf{Labels}} & \textbf{Train/Test Samples} & \multicolumn{1}{c}{\textbf{Website}} \\
        \midrule
        CIFAR-10 & airplane, automobile, bird, cat, deer, dog, frog, horse, ship, truck & 50,000 / 10,000 & \url{http://www.cs.toronto.edu/~kriz/cifar} \\ \midrule
        Fashion-MNIST & T-shirt/top, Trouser, Pullover, Dress, Coat, Sandal, Shirt, Sneaker, Bag, Ankle boot & 60,000 / 10,000 & \url{https://github.com/zalandoresearch/fashion-mnist} \\ \midrule

    \end{tabular}
    \caption{Summary of CIFAR-10 and Fashion-MNIST datasets}
    \label{tab: datasets}
\end{table}

\section{Metrics}\label{metric}
\paragraph{Accuracy.}
In a multi-class classification problem, accuracy is computed using the following formula:

\begin{equation}
    \text{Accuracy} = \frac{\text{Number of correctly predicted samples}}{\text{Total number of samples}}
\end{equation}

Mathematically, accuracy can be expressed as:

\begin{equation}
\text{Accuracy} = \frac{\sum_{i=1}^{N} I(y_i = \hat{y}_i)}{N}
\end{equation}

where:
\begin{itemize}
    \item \(N\) denotes the total number of samples in the test set.
    \item \(y_i\) represents the true class label of the \(i\)-th sample.
    \item \(\hat{y}_i\) is the predicted class label of the \(i\)-th sample by the model.
    \item \(I\) is the indicator function, which equals 1 when \(y_i = \hat{y}_i\) and 0 otherwise.
\end{itemize}

\section{Baselines}\label{BASE}
We perform experiments using various ICL and VICL methods, including Zero-shot Prompt~\cite{brown2020language}, which is without  in-context demonstrations; Static, where the top-k demonstrations are selected from the retrieval database; Random, where demonstrations are randomly chosen for each test input from the retrieval database; Clustering-retrieval~\cite{li2023mot} organizes all demonstrations into 
$k$ clusters, aiming to group similar examples, and subsequently chooses the most representative demonstration from each cluster, resulting in a final set of 
$k$ demonstrations; Kate~\cite{liu2022makes} identifies the most similar examples based on their sentence-level embeddings; and Cluster-Diversity~\cite{naik2023diversity}, where all demonstrations are clustered into 
$k$ groups, and the demonstration closest to each cluster's center is selected to serve as the context demonstration.
VICL-Rerank uses a 'Retrieval \& Rerank' approach by first selecting the top-$k$  most similar samples and then re-ranking them based on the image-text matching score with the query caption to improve relevance.

\begin{table}[t]
\centering
\caption{The  performance of IICL in textual modality.}
\label{Textual_modality_table}
\begin{tabular}{@{}c|cc@{}}
\toprule
                  & Emotion & SST5 \\ \midrule \midrule
Cluster-diversity & 45.0    & 35.7 \\
Random            & 46.5    & 39.4 \\
Cluster-retrieval & 50.1    & 36.5 \\
Static            & 43.5    & 37.9 \\
Kate              & 59.4    & 43.3 \\ \midrule
IJIP              & \textbf{62.3}    & \textbf{45.7} \\ \bottomrule
\end{tabular}
\end{table}

\begin{table}[htbp]
    \centering
    \caption{Performance of IJIP using InternVL 2.5-4B and InternVL 2.5-8B on CIFAR-10 and Fashion-MNIST under varying proportions of missing labels.
}
    \label{missing_label_numbers}
    \begin{tabular}{c c c c c}
        \toprule
        \multirow{2}{*}{\textbf{Class Nums}} & \multicolumn{2}{c}{\textbf{CIFAR10}} & \multicolumn{2}{c}{\textbf{Fashion MNIST}} \\
        \cmidrule(lr){2-3} \cmidrule(lr){4-5}
         & \textbf{4B} & \textbf{8B} & \textbf{4B} & \textbf{8B} \\
        \midrule \midrule
        90\% & 88.5 & 89.9 & 47.5 & 52.8 \\
        80\% & 88.5 & 88.5 & 48.0 & 55.2 \\
        70\% & 88.2 & 88.6 & 50.0 & 59.0 \\
        60\% & 89.4 & 89.0 & 52.4 & 61.0 \\
        50\% & 87.8 & 88.3 & 56.4 & 63.8 \\
        40\% & 88.8 & 91.5 & 64.1 & 68.9 \\
        30\% & 90.5 & 90.8 & 67.9 & 72.7 \\
        20\% & 90.0 & 92.6 & 71.9 & 72.2 \\
        10\% & 92.1 & 92.3 & 77.4 & 78.9 \\
        \bottomrule
    \end{tabular}
\end{table}

\begin{table}[t]
\centering
\caption{The  performance of IJIP with different  LVLMs' sizes. We use InternVL 2.5-2B, InternVL 2.5-4B, InternVL 2.5-8B, and InternVL 2.5-26B.
We evaluate performance primarily using accuracy(\%)↑ under varying proportions of missing labels. The experiments are conducted with label missingness levels of \(10\%\), \(40\%\), and \(90\%\).}
\label{LVLMs_size}
\begin{tabular}{@{}c|ccc|c@{}}
\toprule
LVLMs & 90\% & 40\% & 10\% & Average \\ \midrule \midrule
2B    & 87.2 & 87.5 & 90.6 & 88.4    \\
4B    & 88.5 & 88.8 & 93.9 & 90.4    \\
8B    & 89.9 & 91.5 & 92.3 & 91.2    \\
26B   & 95.4 & 98.1 & 98.3 & 97.3    \\ \bottomrule
\end{tabular}
\end{table}

\begin{table}[t]
\centering
\caption{The IJIP's performances of complete VICL.
The retrieval database includes data corresponding to all labels.
We evaluate performance primarily using accuracy(\%)↑. }
\label{Complete}
\begin{tabular}{@{}c|cc|cc@{}}
\toprule
\multirow{2}{*}{Methods} & \multicolumn{2}{c|}{CIFAR-10 } & \multicolumn{2}{c}{Fashion-MNIST} \\ \cmidrule(l){2-5} 
                         & 8B            & 4B           & 8B              & 4B              \\ \midrule \midrule
Cluster-diversity        & 89.5          & 86.1         & 51.7            & 49.3            \\
Random                   & 89.0          & 83.9         & 53.8            & 54.1            \\
Cluster-retrieval        & 92.0          & 90.5         & 52.7            & 61.1            \\
Static                   & 89.0          & 82.6         & 52.6            & 53.3            \\
Vicl-rerank              & 93.3          & 91.4         & 76.5            & 73.5            \\
Kate                     & 93.7          & 91.9         & 75.7            & 73.5            \\ \midrule
\textbf{IJIP}\cellcolor{lightpurple}& \textbf{94.6}\cellcolor{lightpurple}& \textbf{94.3}\cellcolor{lightpurple}& \textbf{80.9}\cellcolor{lightpurple}& \textbf{80.1}\cellcolor{lightpurple}\\ \bottomrule
\end{tabular}
\end{table}

\end{document}